\documentclass[11pt]{article}

\usepackage[final]{acl}

\usepackage{times}
\usepackage{latexsym}

\usepackage[T1]{fontenc}

\usepackage[utf8]{inputenc}

\usepackage{microtype}
\usepackage{hyperref}
\usepackage{url}
\usepackage{booktabs}

\usepackage{inconsolata}

\usepackage{graphicx}
\usepackage{subcaption}
\usepackage{tcolorbox}
\usepackage{multirow}

\usepackage{amsmath}
\usepackage{amssymb}
\usepackage{mathtools}
\usepackage{amsthm}
\usepackage{xcolor}

\newcommand{\prefilltok}{\fcolorbox{blue!50!black}{blue!18}{\rule{0pt}{0.8ex}\hspace{1.2em}}}
\newcommand{\decodetok}{\fcolorbox{green!40!black}{green!18}{\rule{0pt}{0.8ex}\hspace{1.2em}}}

\newcommand{\best}[1]{\textbf{#1}}

\title{OUTLETS: Output-Length Prediction from Speculative Decoding Backbones}

\author{
  \textbf{Weihuang Wen\textsuperscript{1}},
  \textbf{Yingying Liu\textsuperscript{2}},
  \textbf{Yichuan Liu\textsuperscript{3}},
  \textbf{Wenqi Zeng\textsuperscript{3}},
  \\
  \textbf{Li Zhou\textsuperscript{3}},
  \textbf{Chumin Sun\textsuperscript{3}},
  \textbf{Jie Sun\textsuperscript{3*}},
  \textbf{Tianshu Yu\textsuperscript{1*}}
  \\
  \\
  \textsuperscript{1}The Chinese University of Hong Kong, Shenzhen \\
  \textsuperscript{2}The University of Hong Kong
  \textsuperscript{3}Huawei Technologies Co., Ltd.
  \\
  \small
  \texttt{weihuangwen1@link.cuhk.edu.cn},
  \texttt{u3009796@connect.hku.hk}
  \\
  \small
  \texttt{\{liuyichuan3,zeng.wenqi,zhouli107,sunchumin,j.sun\}@huawei.com}
  \\
  \small
  \texttt{yutianshu@cuhk.edu.cn}
}

\begin{document}
\maketitle
\begingroup
\renewcommand{\thefootnote}{*}
\footnotetext{Corresponding authors.}
\endgroup
\begin{abstract}
The heavy-tailed distribution of output lengths in Large Language Model (LLM) serving poses major challenges for resource provisioning and cluster scheduling. Although output-length prediction can mitigate these issues, existing approaches have key drawbacks: external proxy models add substantial latency and often have limited fidelity, whereas internal state-based methods are efficient but rely on shallow probes of current model states. We identify a structural connection between speculative decoding (SD) and length prediction: latent representations produced by the draft decoder in advanced frameworks (e.g., \mbox{EAGLE-3}) encode signals that are predictive of generation length. Building on this insight, we introduce \textbf{OUTLETS} (\textbf{Out}put-\textbf{Le}ng\textbf{t}h Prediction from \textbf{S}peculative Decoding Backbones), which repurposes the speculative backbone as a trajectory-aware length predictor. When its draft representations are already computed for speculative decoding, OUTLETS adds only a lightweight regression head and achieves lower MAE than the evaluated methods. Under saturated disaggregated serving, OUTLETS predictions enable standard scheduling policies to prioritize shorter requests and distribute requests more evenly across decoding instances, reducing short-request P99 latency by \textbf{34.8\%}.
\end{abstract}

\section{Introduction}
\label{sec:intro}

\begin{figure*}[ht]
    \vskip 0.2in
    \begin{center}
        \begin{subfigure}[b]{0.48\textwidth}
            \centering
            \includegraphics[width=\textwidth]{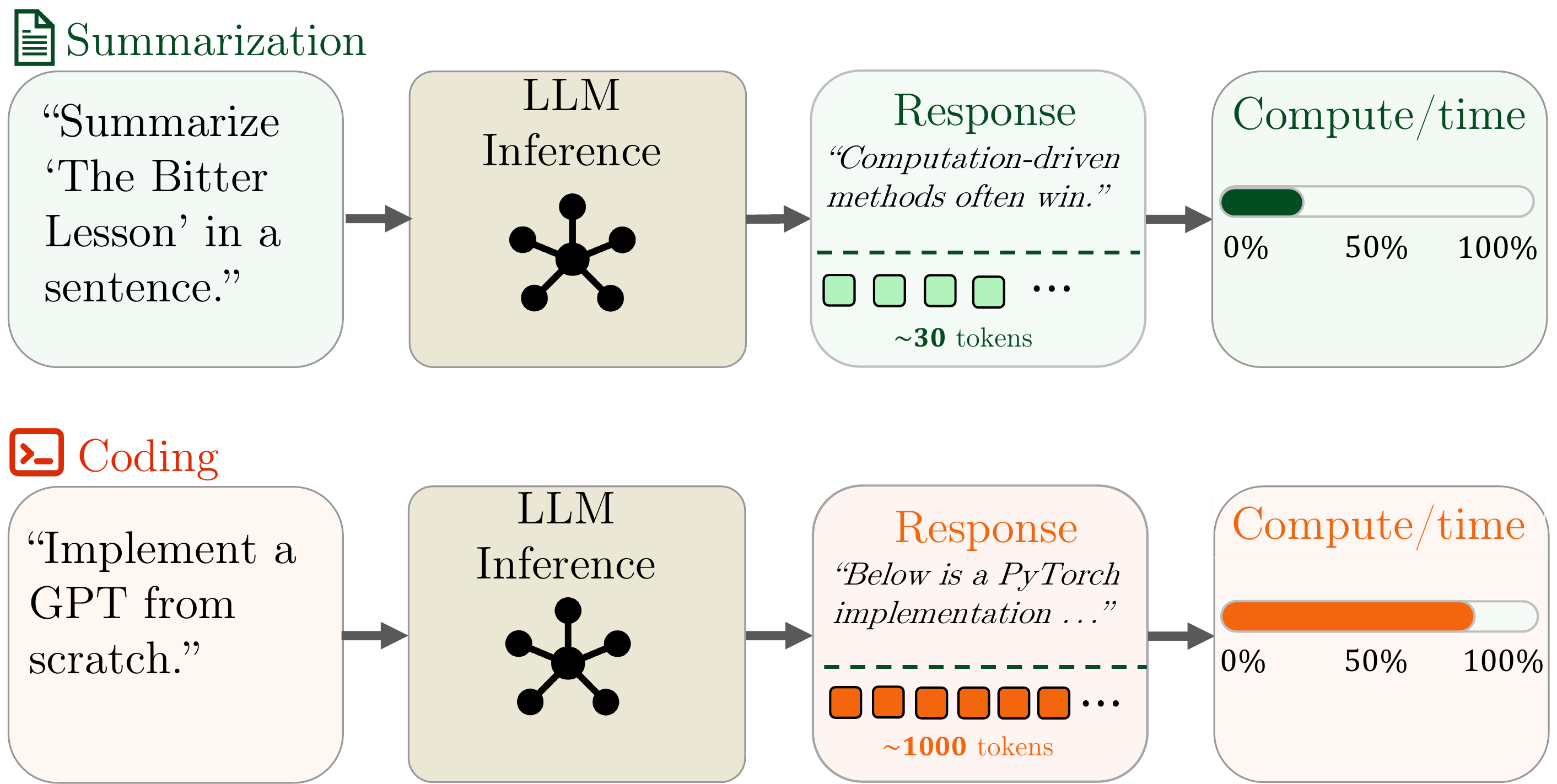}
            \caption{\textbf{Workload Variability.} Distinct application scenarios (e.g., summarization vs. coding) exert drastically different pressures on GPU memory and computational load.}
            \label{fig:1a}
        \end{subfigure}
        \hfill
        \begin{subfigure}[b]{0.48\textwidth}
            \centering
            \includegraphics[width=\textwidth]{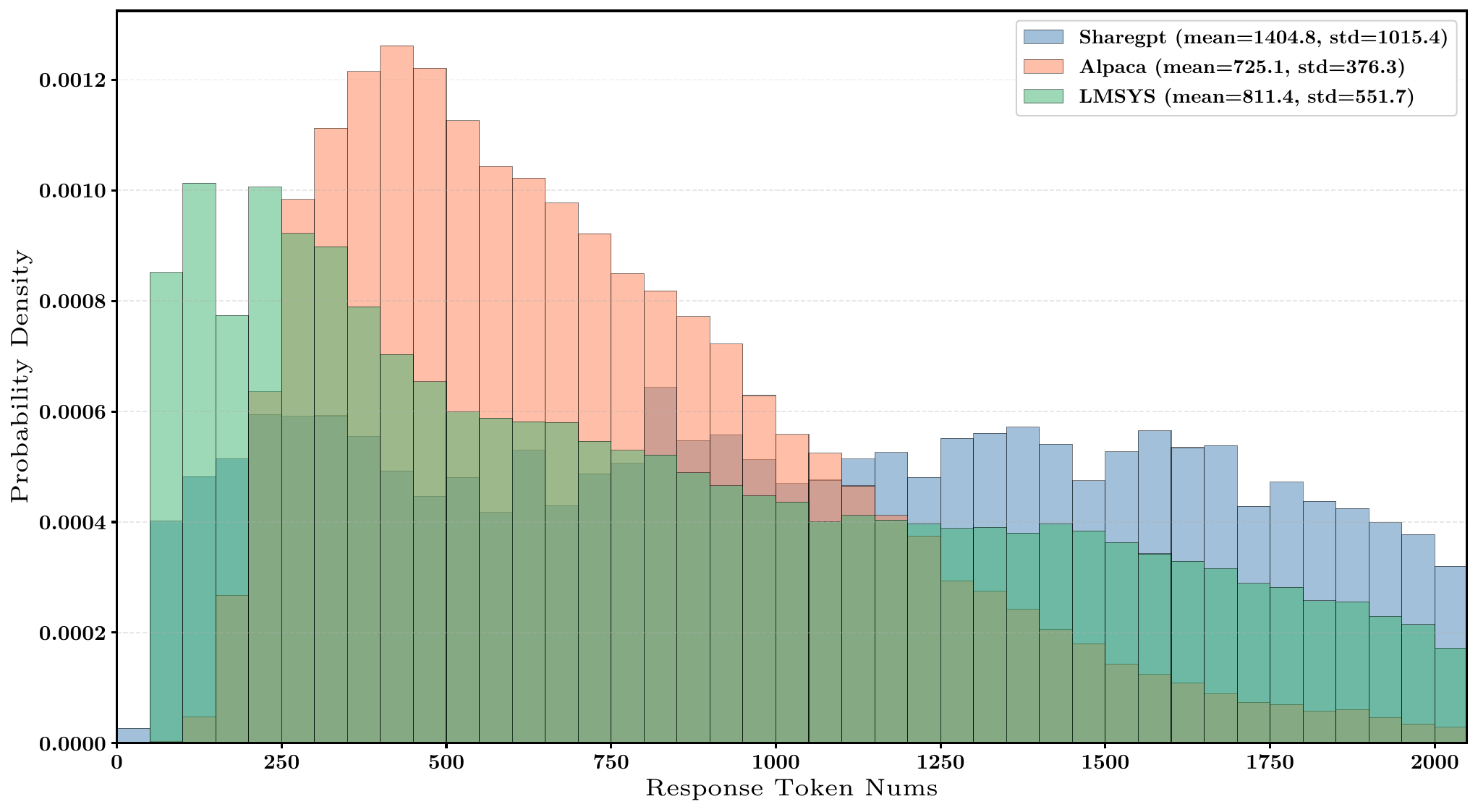}
            \caption{\textbf{Output Length Distributions.} Empirical analysis of Qwen3-30B-A3B outputs reveals length variance spanning orders of magnitude.}
            \label{fig:1b}
        \end{subfigure}
        \caption{\textbf{The challenge of variance in LLM serving.} High stochasticity in output length degrades scheduler efficiency.}
        \label{fig:main_figure}
    \end{center}
    \vskip -0.2in
\end{figure*}

The efficiency of Large Language Model (LLM) serving is fundamentally constrained by the stochastic nature of autoregressive generation ~\cite{zheng2026scheduling, wang2026prod}.
To improve hardware utilization, modern deployment systems increasingly adopt \textbf{disaggregated serving architectures}~\cite{distserve, splitwise}, which decouple prefill and decoding to support massive concurrency and large batch sizes.
Unlike traditional deep learning workloads with deterministic tensor shapes, however, LLM inference exhibits substantial variance.
As illustrated in \textbf{Figure~\ref{fig:1a}}, different user requests impose highly heterogeneous computational loads, while \textbf{Figure~\ref{fig:1b}} shows that output lengths can vary by orders of magnitude even within the same model family.
This variability creates severe system-level bottlenecks when concurrency slots or KV-cache capacity become saturated.
In particular, standard First-Come-First-Served (FCFS) schedulers suffer from Head-of-Line (HOL) blocking, where short, latency-sensitive requests are delayed behind unexpectedly long generations, severely degrading tail latency~\cite{ltr2024}.
At the same time, output-length uncertainty makes agnostic dispatching policies such as round-robin ineffective: without trajectory awareness, they cannot balance load across decoding instances, leading to resource fragmentation and reduced throughput.
As a result, effective scheduling and resource provisioning depend critically on the ability to anticipate generation trajectories.

To address this challenge, \textit{output length prediction} has emerged as a key primitive for advanced scheduling policies.
Yet existing methods face a persistent trade-off between overhead and accuracy.
Proxy-based approaches, such as external BERT regressors, often incur substantial latency and memory overhead while offering limited predictive fidelity.
More recent internal state-based methods improve efficiency by attaching lightweight Multilayer Perceptron (MLP) to the LLM hidden states.
However, these predictors remain shallow and therefore cannot fully exploit the rich semantic and structural information already present in the model representations, especially the long-range signals associated with generation dynamics.

A parallel line of research reduces latency through \textit{speculative decoding} (SD).
SD accelerates generation by using a drafter to propose future tokens, which are then verified by the target model.
We observe a previously underexplored connection between SD and length prediction: both depend on modeling the future evolution of a sequence.
In advanced SD frameworks such as EAGLE-3~\cite{li2025eagle}, the draft decoder constructs hierarchical lookahead features to model upcoming token trajectories.
This differs from conventional hidden-state probing: a probe attached to the target LLM observes states optimized for immediate next-token prediction, whereas a speculative draft decoder is trained to roll representations forward.
Since termination is a property of the future continuation, these lookahead states provide a natural substrate for length prediction.

Motivated by this observation, we propose \textbf{OUTLETS} (\textbf{Out}put-\textbf{Le}ng\textbf{t}h Prediction from \textbf{S}peculative Decoding Backbones), a unified framework with a shared speculative backbone.
OUTLETS connects speculative decoding and output-length prediction at the representation level.
Our goal is not to introduce a new scheduling policy, but to study whether speculative lookahead features can provide the missing length signal needed by existing scheduling mechanisms.
This design repurposes the speculative backbone as a trajectory-aware predictor and reuses computation from the shared drafting graph; when the speculative backbone is already available, the additional cost is dominated by a lightweight regression head.
We further show that the resulting predictions provide an effective scheduling signal in a disaggregated serving system.

Our contributions are summarized as three empirically validated claims:
\begin{enumerate}
    \item \textbf{Speculative lookahead states provide useful length-prediction features.} We reveal a structural connection between speculative decoding and length prediction, and show that draft-decoder representations overcome the expressivity limitations of simple MLP probes over target-model hidden states.
    \item \textbf{Length prediction is compatible with draft-model training.} OUTLETS uses a dual-head framework to jointly optimize token drafting and length regression, and ablations show that the auxiliary objective preserves speculative acceptance while retaining strong static prediction accuracy.
    \item \textbf{The resulting predictions are useful scheduling signals.} We integrate OUTLETS-derived predictions into a real disaggregated serving system, where standard load balancing and Shortest-Job-First scheduling reduce short-request P99 latency and improve throughput.
\end{enumerate}

\section{Related Work}
\label{sec:related_work}
\subsection{Speculative Decoding}
Autoregressive Large Language Model (LLM) inference generates tokens sequentially, activating the full model at every step. This creates substantial latency, especially for long contexts and real-time workloads. Speculative decoding mitigates this bottleneck by letting a draft module propose multiple candidate tokens that are then verified in parallel by the target model.

\textbf{Draft-and-Verify Architectures.}
\citet{xia2023speculative} and \citet{leviathan2023} introduced the ``Draft-and-Verify'' paradigm, where a smaller auxiliary model generates candidate tokens for concurrent verification. Later methods such as DistillSpec~\cite{zhou2024distillspec} improve draft quality through distillation. However, maintaining a separate drafter incurs nontrivial memory and compute costs, and in distributed serving, synchronizing the drafter with a sharded target model can introduce communication overhead that offsets the speedup. Integrated-head designs~\cite{cai2024medusa, ankner2024hydra} provide another direction, but are orthogonal to our setting.

\textbf{The EAGLE Family.}
EAGLE-1~\cite{li2024eagle} performs drafting in continuous feature space rather than directly predicting tokens. It uses hidden states, particularly from the second-to-top layer, together with time-shifted tokens to better model future states under feature uncertainty. EAGLE-2~\cite{li2024eagle2} further observes that draft acceptance depends on context as well as position, and proposes Context-Aware Dynamic Draft Trees to adapt the draft structure using confidence scores. EAGLE-3~\cite{li2025eagle} addresses the saturation issues of earlier versions by switching to direct token prediction with feature fusion, and introduces Training-Time Test (TTT) to reduce the train--test gap and stabilize multi-step acceptance.

\textbf{Adaptive Speculation Length.}
Recent works also adapt the number of candidate tokens drafted within each
speculative-decoding iteration.  DISCO dynamically selects the speculation
lookahead instead of using a fixed value, while SpecDec++ learns an acceptance
prediction head and stops drafting when the predicted rejection probability
crosses a threshold \citep{mamou2024disco,huang2025specdecpp}.  These methods
optimize token-level draft length within a speculative iteration.

\subsection{Output Length Prediction}
\textbf{Instruction-Based Methods.}
Early methods leverage the semantic capability of the LLM itself. \citet{pia2023} propose \textit{Perception in Advance} (PIA), which instruction-tunes the model to explicitly predict output length. This can support micro-batching and reduce padding overhead, but it requires prompt modification and weight updates that may affect alignment. It also adds latency because prediction tokens must be generated before the actual response.

\textbf{Proxy-Based Methods.}
Another line of work decouples prediction from serving by using lightweight auxiliary models. $S^3$~\cite{s32023} uses a DistilBERT regressor for memory planning, while \citet{ssjf2024} extend this idea to \textit{Speculative Shortest-Job-First} scheduling with a BERT proxy. \citet{shuffleinfer} use a small LLM to bucket output lengths, and \citet{ltr2024} adopt a similar predictor within a \textit{Learning to Rank} framework. These approaches are flexible, but they rely on external models that add system complexity and consume extra resources. Their predictors also lack direct access to the target model's architecture and runtime states, which can limit fidelity.

\textbf{Internal State-Based Methods.}
Recent work moves toward model-native prediction by probing internal activations. TRAIL~\cite{shahout2025trail} supports Shortest Remaining Processing Time (SRPT) scheduling by feeding hidden states into a lightweight classifier for continuous prediction. ARES~\cite{wang2025ares} similarly estimates remaining generation time from hidden states to mitigate workload imbalance in disaggregated clusters. For reasoning models, \citet{anonymous2025overclocking} identify hidden states correlated with chain-of-thought progress and show that these states can be manipulated to remove redundant reasoning steps. More recently, \citet{xie2026egtp} combine internal activations with token entropy for low-cost prediction.

\subsection{Connection to Our Work}
We identify a structural connection between speculative decoding and output-length prediction. Advanced SD methods such as EAGLE produce rich representations that simulate future generation trajectories, yet these signals are typically used only for token drafting. Existing internal predictors such as TRAIL usually attach lightweight MLP heads to the target LLM, whose states are primarily optimized for next-token prediction at the current step. OUTLETS bridges these two lines of work by repurposing the speculative backbone itself for length prediction. Since the draft decoder is trained to model upcoming continuations, its lookahead states offer a more trajectory-aware substrate than standalone probes over frozen hidden states. This design combines the efficiency of model-native predictors with the richer predictive capacity of advanced speculative backbones, while adding only a small marginal cost when the speculative backbone is already present.

\begin{figure*}[t]
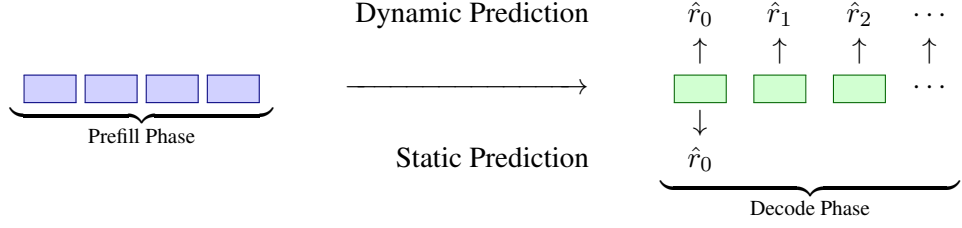

\centering
\[
\begin{array}{c@{\qquad}c@{\qquad}c}
\underbrace{
\begin{array}{c}
\prefilltok\;\prefilltok\;\prefilltok\;\prefilltok
\end{array}
}_{\text{Prefill Phase}}
&
\begin{array}{r}
\text{Dynamic Prediction} \\
\\
\xrightarrow{\hspace{3cm}} \\
\\
\text{Static Prediction}
\end{array}
&
\underbrace{
\begin{array}{cccc}
\hat{r}_0 & \hat{r}_1 & \hat{r}_2 & \cdots \\
\uparrow & \uparrow & \uparrow & \uparrow \\
\decodetok & \decodetok & \decodetok & \cdots \\
\downarrow & & & \\
\hat{r}_0 & & &
\end{array}
}_{\text{Decode Phase}}
\end{array}
\]
\caption{Static prediction is performed once after prefill, while dynamic prediction is updated throughout decoding as new tokens are generated.}
\label{fig:static_dynamic_prediction}
\end{figure*}

\section{Preliminary}
\label{sec:preliminary}

We formalize output-length prediction in LLM serving, with the goal of estimating the \emph{remaining} number of tokens during generation. This leads to two practical settings: \emph{static} prediction before decoding starts, and \emph{dynamic} prediction during decoding.

\subsection{Problem Formulation}

Consider an inference request with input prompt $\mathbf{x} = (x_1, \dots, x_n)$. Conditioned on $\mathbf{x}$, the LLM autoregressively generates an output sequence $\mathbf{y} = (y_1, \dots, y_L)$, where $L$ is the total number of generated tokens. Generation stops once a stopping criterion is met, such as producing an EOS token.

At decoding step $t$, where $0 \le t < L$, let $\mathbf{h}_t$ denote the model state available at that step, e.g., the hidden representation from the final Transformer layer. We define the \textbf{true remaining length} as $r_t = L - t$, and seek a predictor $f_\theta$ that estimates $r_t$ from the information available at step $t$.

\subsection{Static vs. Dynamic Prediction}

\textbf{Static remaining length prediction.}
Static prediction is performed at step $t = 0$, after the prompt has been processed in the prefill phase and before any output token is generated. In this case, the remaining length equals the total output length, i.e., $r_0 = L$, and the predictor estimates $\hat{r}_0 = f_\theta(\mathbf{x}, \mathbf{h}_0)$, where $\mathbf{h}_0$ summarizes the prompt context after prefill.
Static prediction is useful for initial resource planning, such as instance-level load balancing in disaggregated serving. However, because it relies only on prompt-side information, it cannot reflect the realized generation trajectory and may be inaccurate when output lengths vary substantially.

\textbf{Dynamic remaining length prediction.}
Dynamic prediction is performed during decoding, i.e., at steps $t \ge 1$. As generation unfolds, the predictor continually updates its estimate using both the prompt and the partially generated response. Formally, at step $t$, it predicts $\hat{r}_t = f_\theta(\mathbf{x}, \mathbf{y}_{1:t}, \mathbf{h}_t)$, where $\mathbf{y}_{1:t} = (y_1, \dots, y_t)$ denotes the generated prefix up to step $t$.
Because dynamic prediction conditions on the realized decoding trajectory, it can refine its estimate as more evidence becomes available. In serving systems, this enables online estimation of the projected total generation length, $t + \hat{r}_t$, allowing the scheduler to react when a request runs longer than expected or finishes earlier than predicted. Figure~\ref{fig:static_dynamic_prediction} illustrates the difference between static and dynamic prediction.

\section{Method}
\label{sec:method}

\begin{figure*}[t]
  \centering
  \includegraphics[width=\textwidth]{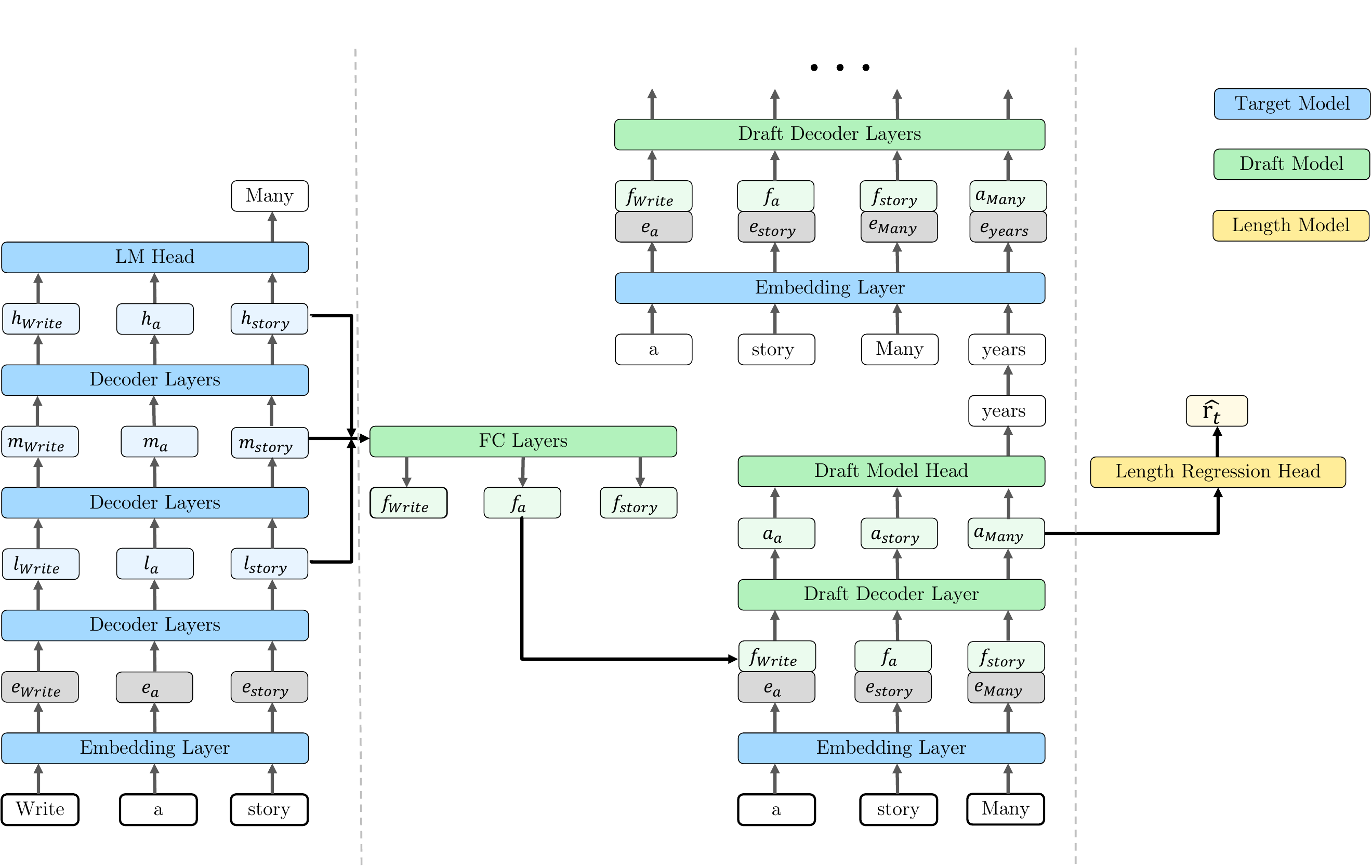}
  \caption{\textbf{Overview of the OUTLETS architecture.} OUTLETS unifies speculative drafting and sequence length regression within a shared computational graph. (Left) During inference, multi-level hidden states ($l, m, h$) are extracted and fused through an FC layer to capture rich semantic signals. (Right) The shared speculative backbone processes these fused features to jointly perform two tasks: (1) generating draft tokens through the Draft Model Head for speculative decoding, and (2) predicting output length through a lightweight Length Regression Head.}
  \label{fig:framework}
\end{figure*}

In this section, we present OUTLETS, a unified framework for jointly optimizing speculative decoding and output length prediction. Figure~\ref{fig:framework} provides an overview. Our design is motivated by the hypothesis that the latent states of an expressive speculative drafter encode not only local token-level dynamics, but also higher-level structural signals related to sequence termination. OUTLETS disentangles these signals with a dual-head architecture that enables accurate length prediction with minimal additional cost.

\subsection{Architectural Backbone}
\label{subsec:backbone}

OUTLETS follows the architectural philosophy of the EAGLE family~\cite{li2024eagle, li2024eagle2, li2025eagle}, while addressing the limitations of naive speculative decoding through a dedicated draft decoder. The backbone contains two main components: a feature fusion layer and a draft decoder.

\textbf{Feature Fusion Layer.}
Many speculative decoding methods rely only on the top-layer hidden states of the target LLM. These representations are often highly specialized for next-token prediction and may be less suitable for longer-horizon planning. Following~\cite{li2025eagle}, we therefore fuse hidden states from multiple depths to recover richer contextual signals for both drafting and length prediction.
Let $\mathcal{H} = \{\mathbf{h}^{(i)} \mid i \in [0, N]\}$ denote the hidden states of the frozen target LLM with $N$ layers. We select a subset of layers $\mathcal{I} = \{2, \lfloor N/2 \rfloor, N-2\}$, corresponding roughly to shallow lexical features, intermediate syntactic patterns, and deeper semantic abstractions. These representations are concatenated and projected into the draft embedding space $\mathbb{R}^d$:
\begin{equation}
\mathbf{f}_{\text{ctx}} = \mathbf{W}_{\text{proj}} \left( \text{Concat}\left(\{\mathbf{h}^{(i)}\}_{i \in \mathcal{I}}\right) \right).
\end{equation}
The fused feature $\mathbf{f}_{\text{ctx}}$ serves as a static contextual anchor for autoregressive generation.

\textbf{Draft Decoder Layer.}
To capture temporal dynamics, we use a lightweight Transformer decoder $\mathcal{M}_{\text{draft}}$ with self-attention and MLP blocks. We instantiate attention with \textbf{Gated Attention}~\cite{qiu2025gated} rather than the standard attention used in EAGLE-3, because it mitigates attention sink and better preserves long-range termination signals. Our experiments show consistent gains in length-prediction accuracy over standard Llama attention.

\subsection{Dual-Head Formulation}
\label{subsec:dual_head}

To extract both lexical and structural information from the shared draft state, we use a dual-head design that maps the same latent representation to two task-specific outputs.

\textbf{Draft Model Head.}
The first branch follows the standard speculative decoding setup. It maps the draft decoder state $\mathbf{h}_{t}^{\text{draft}}$ at step $t$ to the vocabulary space $\mathcal{V}$ to approximate the target model's next-token distribution:
\begin{equation}
\mathbf{P}_{\text{draft}}(y_{t+1} \mid y_{1:t}) = \text{softmax}\left(\mathbf{W}_{\text{lm}} \cdot \mathbf{h}_{t}^{\text{draft}}\right).
\end{equation}

\textbf{Length Regression Head.}
The second branch probes the same draft state for a global structural signal: the remaining sequence length. Since generation lengths are typically heavy-tailed, direct prediction in linear space can be unstable. We therefore regress in log-space and define the target as $z = \log(1 + \text{length})$.
The regression head is implemented as an MLP with progressively reduced hidden dimensions:
$\hat{z}_t = \text{MLP}_{\text{len}}\left(\mathbf{h}_{t}^{\text{draft}}\right)$.
Here, $\text{MLP}_{\text{len}}$ consists of linear layers interleaved with ReLU activations. The scalar output $\hat{z}_t$ represents the model's estimate of the remaining generation cost at step $t$.

\textbf{Joint Training Objective.}
We train the model end-to-end with a composite objective that combines local generation quality with global structure prediction:
\begin{equation}
\mathcal{L} = \mathcal{L}_{\text{SD}} + \lambda \mathcal{L}_{\text{len}} + \gamma \|\theta_{\text{MLP}}\|_2^2.
\end{equation}
Here, $\mathcal{L}_{\text{SD}}$ denotes the speculative decoding KL divergence, and $\mathcal{L}_{\text{len}}$ is the log-space mean squared error for length prediction. The last term applies $L_2$ regularization to the length-prediction MLP parameters $\theta_{\text{MLP}}$ to improve stability and reduce overfitting; we set $\gamma = 10^{-5}$.

To balance the two tasks, we also evaluated adaptive weighting methods such as Uncertainty Weighting~\cite{kendall2018multi} and GradNorm~\cite{chen2018gradnorm}. In practice, however, a simple fixed weight gave the best trade-off between stability and performance. We therefore set $\lambda = 0.1$, which provides sufficient structural supervision without interfering with the primary drafting objective.

\begin{table*}[t]
\caption{Performance comparison of length prediction methods. We report the mean absolute error (MAE), where lower values indicate better accuracy. OUTLETS consistently achieves the lowest error rates across all model architectures and scenarios. ``Avg. Len'' denotes the total output-length statistics for static prediction and the remaining-length target statistics aggregated over decoding steps for dynamic prediction ($\mu \pm \sigma$).}
\label{tab:scenario-comparison}
\centering
\small
\setlength{\tabcolsep}{3.5pt}
\renewcommand{\arraystretch}{1.05}
\begin{tabular}{@{}lll c ccccc@{}}
\toprule
 & & & & \multicolumn{5}{c}{Method} \\
\cmidrule(lr){5-9}
Model & Scenario & Dataset & Avg. Len & OUTLETS & MLP & BERT & OPT & PIA \\
\midrule
\multirow{6}{*}{Llama-3.2-1B} & \multirow{3}{*}{Static}
  & ShareGPT & 324.4 $\pm$ 268.9 & \best{100.1} & 109.5 & 128.6 & 130.6 & 204.3 \\
& & Alpaca   & 237.6 $\pm$ 198.3 & \best{53.4}  & 56.9  & 70.0  & 70.4  & 132.7 \\
& & LMSYS    & 246.6 $\pm$ 276.8 & \best{69.6}  & 73.6  & 83.7  & 84.5  & 171.8 \\
\cmidrule(l){2-9}
& \multirow{3}{*}{Dynamic}
  & ShareGPT & 274.2 $\pm$ 226.7 & \best{63.8} & 72.5 & - & - & - \\
& & Alpaca   & 202.1 $\pm$ 175.8 & \best{31.8} & 39.1 & - & - & - \\
& & LMSYS    & 279.1 $\pm$ 256.7 & \best{41.7} & 50.3 & - & - & - \\
\midrule
\multirow{6}{*}{Llama-3.1-8B} & \multirow{3}{*}{Static}
  & ShareGPT & 355.9 $\pm$ 276.7 & \best{80.6} & 84.9 & 132.8 & 135.2 & 283.3 \\
& & Alpaca   & 250.3 $\pm$ 206.9 & \best{46.0} & 46.8 & 102.9 & 112.8 & 285.1 \\
& & LMSYS    & 257.6 $\pm$ 299.0 & \best{70.6} & 92.4 & 126.7 & 144.3 & 178.6 \\
\cmidrule(l){2-9}
& \multirow{3}{*}{Dynamic}
  & ShareGPT & 286.0 $\pm$ 232.9 & \best{67.6} & 90.4 & - & - & - \\
& & Alpaca   & 211.2 $\pm$ 174.7 & \best{39.4} & 51.2 & - & - & - \\
& & LMSYS    & 302.9 $\pm$ 292.7 & \best{38.8} & 48.0 & - & - & - \\
\midrule
\multirow{6}{*}{Qwen3-30B-A3B} & \multirow{3}{*}{Static}
  & ShareGPT & 1053.9 $\pm$ 536.8 & \best{186.7} & 210.8 & 262.2 & 255.2 & 936.2 \\
& & Alpaca   & 725.9 $\pm$ 374.2  & \best{172.1} & 189.0 & 200.0 & 201.8 & 656.2 \\
& & LMSYS    & 791.7 $\pm$ 536.1  & \best{166.8} & 189.1 & 224.4 & 229.4 & 800.0 \\
\cmidrule(l){2-9}
& \multirow{3}{*}{Dynamic}
  & ShareGPT & 664.1 $\pm$ 460.8 & \best{117.1} & 158.4 & - & - & - \\
& & Alpaca   & 459.9 $\pm$ 353.9 & \best{94.0}  & 117.4 & - & - & - \\
& & LMSYS    & 578.8 $\pm$ 441.2 & \best{103.2} & 130.6 & - & - & - \\
\bottomrule
\end{tabular}
\vskip -0.1in
\end{table*}

\section{Experiments}

\subsection{Experimental Setup}
\label{subsec:exp_setup}

We summarize the experimental setup here and defer full details to Appendix~\ref{app:exp_details}.

\textbf{Implementation.}
We build on the open-source EAGLE-3 and SpecForge~\cite{specforge} codebases and use DeepSpeed for distributed training. All experiments run on a cloud node with four NVIDIA GPUs (141~GB HBM in total) and two 56-core CPUs. Models are trained for up to 10 epochs with scale-dependent batch sizes, and the random seed is fixed to 0.

\textbf{Datasets.}
We use \textbf{ShareGPT}~\cite{sharegpt}, \textbf{Alpaca}~\cite{alpaca}, and \textbf{LMSYS-Chat-1M}~\cite{lmsys}, covering real-world multi-turn dialogue, synthetic single-turn instruction following, and large-scale open-ended conversations. For each target model, we regenerate responses on the same prompts and use the resulting generation lengths as model-specific labels. We filter out samples whose generated length exceeds 2,048 tokens and split the remaining data into training and test sets with a 4:1 ratio. For Qwen3-30B-A3B, we retain only the
first turn of ShareGPT and LMSYS to avoid inconsistent per-turn length
accounting caused by compressed or dropped thinking content, and use a 1/10
subsample of LMSYS due to generation cost. Full dataset statistics are reported
in Appendix~\ref{app:exp_details}.

\textbf{Models.}
We evaluate \textbf{Llama-3.2-1B-Instruct}, \textbf{Llama-3.1-8B-Instruct}, and \textbf{Qwen3-30B-A3B}, covering standard dense models and a reasoning-capable MoE model; details are provided in Appendix~\ref{app:model_details}.

\textbf{Baselines.}
We compare against three families of methods. For internal state-based prediction, we use an \textbf{MLP} regressor attached to intermediate hidden states of the frozen target LLM, following prior model-native predictors such as TRAIL~\cite{shahout2025trail}, ARES~\cite{wang2025ares}, and Overclocking~\cite{anonymous2025overclocking}. For proxy-based methods, we include \textbf{BERT}, a semantic regressor similar to $S^3$~\cite{s32023}, and \textbf{OPT}, a classification-based proxy in the spirit of ShuffleInfer and LTR~\cite{shuffleinfer}. For instruction-based prediction, we include a \textbf{PIA-style prompting} baseline that asks the model to emit a length estimate before generation; we use it as a lightweight instruction-based reference rather than a fully fine-tuned reproduction of PIA~\cite{pia2023}.

\textbf{Metrics.}
We report MAE for both static and dynamic prediction. For static prediction, $\text{MAE}=\frac{1}{N}\sum_{i=1}^{N}\left|L^i-\hat{r}_0^i\right|$, where $L^i$ is the ground-truth output length of request $i$ and $\hat{r}_0^i$ is the prediction at $t=0$. For dynamic prediction, with $\hat{r}_t^i$ denoting the predicted remaining length at decoding step $t$, we report $\text{MAE}=\frac{1}{N}\sum_{i=1}^{N}\left(\frac{1}{L^i}\sum_{t=0}^{L^i-1}\left|(L^i-t)-\hat{r}_t^i\right|\right)$.

\subsection{Results and Analyses}

\textbf{Static Prediction.}
As shown in Table~\ref{tab:scenario-comparison}, OUTLETS consistently outperforms proxy-based methods (e.g., BERT and OPT)
and the PIA-style prompting baseline across all datasets. This supports our hypothesis that hierarchical features derived from internal LLM states provide stronger semantic grounding than the shallow token-level representations used by external proxies.

\textbf{Dynamic Prediction.}
Proxy- and instruction-based methods are impractical for dynamic prediction because per-step querying introduces prohibitive latency. We therefore compare OUTLETS only against the MLP baseline in this setting. OUTLETS' MAE drops steadily as decoding progresses (e.g., from 80.6 to 67.6 on ShareGPT), indicating that the draft state $\mathbf{h}_{t}^{\text{draft}}$ captures trajectory information that supports increasingly accurate updates. In this work, dynamic prediction serves as a representation-level test of whether the shared draft state tracks generation progress; the deployed scheduling policy below uses static predictions because admission and routing decisions must be made before decoding starts.

\textbf{Scaling Properties.}
OUTLETS substantially outperforms the MLP baseline, and its advantage grows in the dynamic setting as training data increases, mirroring the scaling behavior of the EAGLE-3 backbone. Unlike a simple MLP, which lacks the inductive bias to model long-range dependencies, the Transformer-based backbone in OUTLETS continues to benefit from more supervision.

\textbf{Computational Overhead.}
OUTLETS introduces modest standalone overhead, mainly determined by the target LLM hidden size $d$. Under a representative configuration with $d=2,048$ (e.g., Qwen3-30B-A3B), the speculative decoding backbone and prediction MLP add about 72.3M and 2.6M parameters, respectively. In runtime, the SD backbone contributes about 1.7~ms per decoding step, while the prediction head adds only 0.7~ms. This low-overhead characterization applies when the speculative backbone is already available: the regression head is lightweight, while the draft-backbone cost should be counted when OUTLETS is used solely as a standalone predictor.

\subsection{System-Level Evaluation in Disaggregated Serving}
\label{sec:system_eval}

To assess the practical utility of the prediction signal, we integrate OUTLETS into a production-grade disaggregated serving system with one prefill instance and three decode instances, and evaluate it under high-concurrency stress tests on ShareGPT. These scheduling policies are standard control-plane mechanisms; our evaluation focuses on whether OUTLETS can supply an accurate length signal that makes them effective in an online serving stack. Additional workload filtering details and saturation-time breakdowns are provided in Appendix~\ref{app:system_eval}.

\textbf{Operating regime and evaluation scope.}
Speculative acceleration and length-aware scheduling address different serving bottlenecks. In common deployment settings, speculative decoding improves token-level latency by exploiting otherwise idle compute during memory-bound or underutilized decoding \cite{DBLP:conf/aaai/ShiLZQLZ26}, whereas length-aware scheduling is most valuable under saturated high-concurrency serving, where queueing delay, KV-cache pressure, and inter-instance load imbalance dominate tail latency ~\cite{ltr2024}. Our system experiment therefore focuses on this saturated regime. Enabling speculative acceleration would change the effective service rate and resource-contention pattern, making it difficult to attribute improvements to the OUTLETS scheduling signal itself. We thus disable speculative acceleration in all compared configurations, keep the decoding service model fixed, and evaluate OUTLETS as a control-plane prediction signal for routing and queue ordering.

\textbf{System Setup.}
Our system-level study focuses on the decode stage, where scheduling most strongly affects queueing delay under load. We consider two scheduling decisions: \emph{instance routing}, which determines which decode instance receives an incoming request, and \emph{queue scheduling}, which determines the service order within each decode queue. We compare two scheduling frameworks. \textbf{(1) Baseline (vLLM v0.13.0):} a standard vLLM backend with \textbf{Round-Robin (RR)} dispatch and \textbf{First-Come-First-Served (FCFS)} admission, where requests are routed to decode instances in a fixed cyclic order and served in arrival order. \textbf{(2) Ours (Load Balancing + Shortest-Job-First):} a length-aware policy that uses the static OUTLETS prediction $\hat{r}_0$ as a proxy for decode cost in two places. For \textbf{Load Balancing (LB)}, each incoming request is routed to the decode instance with the lowest estimated remaining workload, computed from the predicted remaining lengths of its queued and running requests. For \textbf{Shortest-Job-First (SJF)} scheduling, requests with shorter predicted generations are given higher priority within each decode queue. This design reduces HOL blocking. As the main tail-latency metric, we report \textbf{P99 latency}, i.e., the 99th-percentile end-to-end request latency, because it is closely tied to user-visible responsiveness and service-level objectives (SLOs).

\begin{figure}[t]
    \centering
    \includegraphics[width=\linewidth]{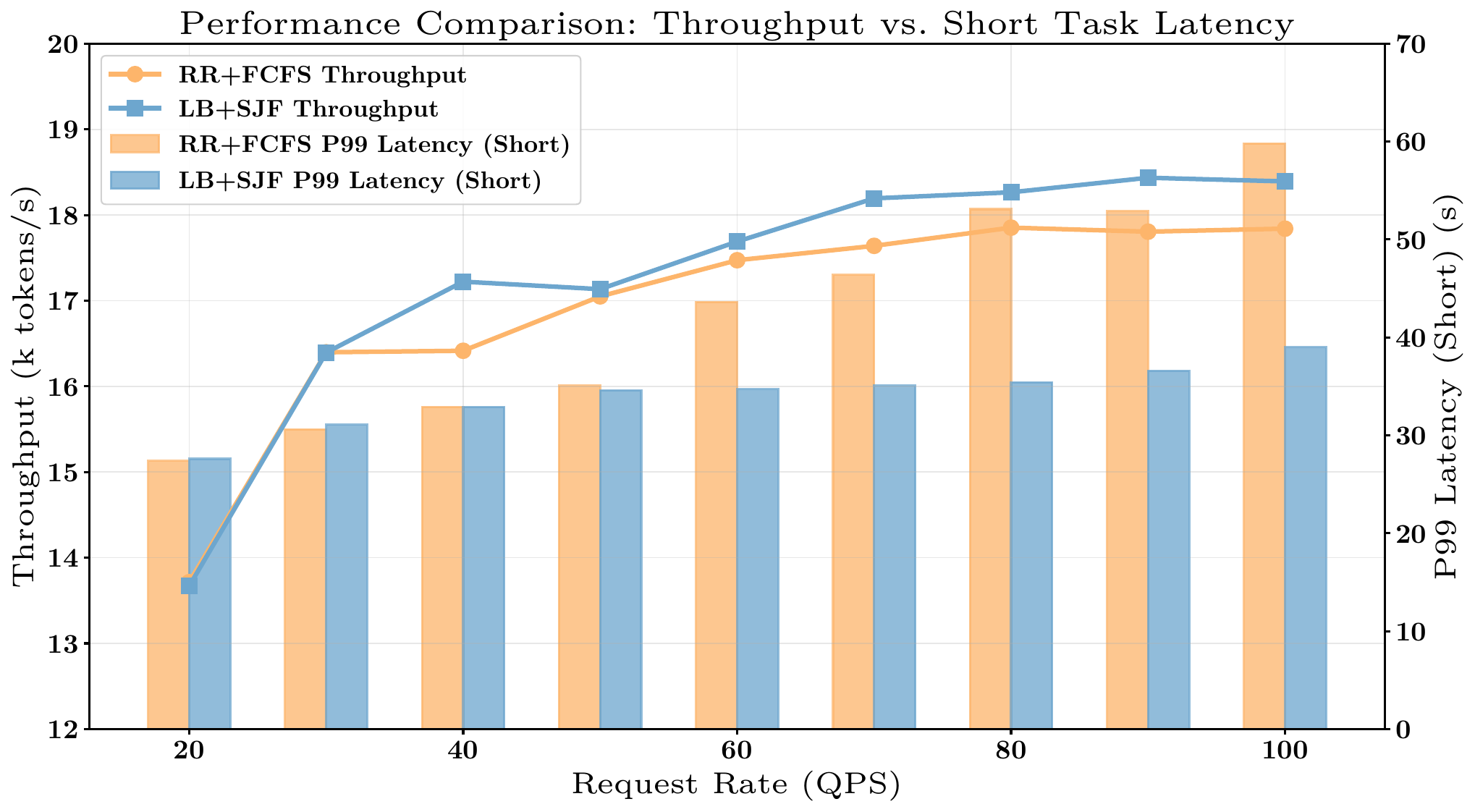}
    \caption{\textbf{End-to-end serving performance.} Comparison of the vLLM baseline (RR + FCFS) and OUTLETS-guided scheduling (LB + SJF). Our method substantially reduces tail latency for short tasks by mitigating Head-of-Line blocking.}
    \label{fig:serving_perf}
\end{figure}

\textbf{System-Level Improvements.}
As shown in Figure~\ref{fig:serving_perf}, replacing the RR+FCFS baseline with OUTLETS-guided LB+SJF improves both throughput and tail latency. At saturation (100 QPS), peak throughput increases from 17.8k to 18.4k tokens/s, while the P99 latency of short requests ($<800$ tokens) decreases by \textbf{34.8\%}, from 59.8~s to 39.0~s. Further analysis in Appendix~\ref{app:system_eval} attributes most of the tail-latency reduction to local SJF scheduling, which mitigates HOL blocking, while instance-level load balancing provides an additional improvement in throughput. We also conduct controlled experiments under both RR+SJF and LB+SJF by holding the scheduling policy fixed and varying only the static output-length predictor, with no-prediction and Oracle references. The results, reported with reproducibility details in Appendix~\ref{draft:predictor-system}, confirm that the OUTLETS predictions provide an effective signal for distributing decode workloads and prioritizing short requests.

\subsection{Ablation Studies}
We conduct ablation studies to address three possible concerns about OUTLETS. First, joint training might harm drafting quality; comparing OUTLETS with an SD-ONLY model shows that adding the length-regression objective does not materially reduce speculative acceptance. Second, a dedicated length predictor might be strictly better; comparing against an LP-ONLY model shows that OUTLETS is generally comparable or better in the static setting, while LP-ONLY has a modest advantage in dynamic prediction where step-level supervision is much denser. Third, the gains might come merely from adding parameters; component ablations show that removing the draft decoder causes the largest degradation, and direct regression consistently outperforms bucket-based classification. These results support our design choice of reusing the speculative backbone as a trajectory-aware forecasting module. Full results are provided in Appendix~\ref{app:ablation}.

\section{Conclusion}
\label{sec:conclusion}

In this paper, we introduced OUTLETS, a framework that repurposes speculative-decoding backbones for output-length prediction. Across the evaluated models and datasets, representations produced by the speculative drafter yield lower prediction MAE than shallow MLP probes over target-model hidden states. When these representations are already computed for speculative decoding, OUTLETS introduces only a lightweight regression head for length prediction. We further demonstrate that the resulting static predictions provide effective control-plane signals in a disaggregated serving system, enabling instance-level load balancing and SJF scheduling to reduce HOL blocking, lower P99 latency for short requests, and improve overall throughput. Taken together, our results show that speculative drafters can support not only token-level acceleration but also higher-level serving decisions.

\section*{Limitations}
\label{sec:limitations}

\paragraph{Deployment cost.}
OUTLETS has the smallest marginal cost when a speculative backbone is already available. If it is used purely as a standalone length predictor, the cost of the draft backbone must be counted, and lighter predictors may be preferable in extremely latency-sensitive settings.

\paragraph{Scheduling scope.}
Our system integration uses static predictions for admission-time load balancing and SJF scheduling. Dynamic predictions are evaluated as evidence that the shared draft state tracks generation progress, but we do not yet use them for online migration or rescheduling. We also do not implement an adaptive runtime that switches between speculative acceleration and saturation-mode scheduling; our system study isolates the latter regime to evaluate the control-plane value of the OUTLETS prediction signal.

\paragraph{Length range and decoding policy.}
Our evaluation is bounded by a maximum output length of 2,048 tokens and a fixed decoding setup. Prediction accuracy under very long agentic trajectories, different sampling temperatures, custom stopping conditions, or direct KV-cache statistics remains to be verified.

\section*{Acknowledgments}
This work was supported in part by the National Key R\&D Program of China under grant 2022YFA1003900.
We would like to thank all the anonymous reviewers for their insightful comments.



\bibliography{custom}

\appendix

\section{Use of Large Language Models (LLMs)}
During the preparation of this manuscript, we used Large Language Models (LLMs) for language assistance, including grammar checking and text polishing, to improve clarity and readability.

In addition, during the experimental phase, we used LLMs to assist in generating functional scripts (e.g., dataset generation, system benchmarking harnesses, and metric collection/statistics scripts). All such code was reviewed, tested, and, where necessary, modified by the authors. The authors take full responsibility for the correctness of the implementation, the experimental design, and the reported results and conclusions. LLMs were not used to generate experimental results or to make scientific decisions.

\section{Experimental Details}
\label{app:exp_details}

\subsection{Hyperparameter Configuration}
We train all models using the AdamW optimizer with learning rate 1e-5, weight decay 0.02, and betas (0.9, 0.95). Training uses 2,000 warmup steps with linear decay. We apply gradient clipping at 1.0 and use BF16 mixed precision.
All sequences are truncated to 2,048 tokens.

\subsection{Dataset Descriptions}

\textbf{Datasets.}
We use three public Hugging Face datasets: ShareGPT~\footnote{\url{https://huggingface.co/datasets/shibing624/sharegpt_gpt4}}, Alpaca~\footnote{\url{https://huggingface.co/datasets/tatsu-lab/alpaca}}, and LMSYS-Chat-1M~\footnote{\url{https://huggingface.co/datasets/lmsys/lmsys-chat-1m}}. ShareGPT and LMSYS consist of multi-turn conversations, whereas Alpaca is single-turn instruction-following data. The original dataset sizes are 58,674 (ShareGPT), 52,000 (Alpaca), and 1,000,000 (LMSYS). For each target model, following the same pipeline used to construct speculative-decoding training data, we re-generate model outputs on the corresponding prompts to obtain consistent ground-truth lengths under that model. We retain only samples whose generated length is $<2,048$ tokens; longer generations are excluded because truncation prevents accurate measurement of the true output length. We then split the cleaned set into training and test partitions with a 4:1 ratio. Since re-generation and length-based filtering are model-dependent, the resulting dataset sizes vary slightly across target models.

For Qwen3-30B-A3B, we observe that in multi-turn chat the model may drop earlier thinking content when forming the context in later turns, making per-turn output-length accounting inconsistent. Therefore, for ShareGPT and LMSYS we keep only the first turn when constructing length labels for this model. In addition, due to the high computational cost of full-scale generation and training for LMSYS on Qwen3-30B-A3B, we use only a 1/10 subsample of LMSYS for this setting. Table~\ref{tab:dataset_statistics} reports the final train/test sizes after re-generation and filtering.

\begin{table}[t]
\centering
\small
\setlength{\tabcolsep}{3.0pt}
\renewcommand{\arraystretch}{1.04}
\caption{Final dataset sizes after response regeneration and length filtering.}
\label{tab:dataset_statistics}
\begin{tabular*}{\columnwidth}{@{\extracolsep{\fill}}llrr@{}}
\toprule
Model & Dataset & Train & Test \\
\midrule
\multirow{3}{*}{Llama-3.2-1B}
  & ShareGPT & 13,649  & 3,363 \\
  & Alpaca   & 41,426  & 10,352 \\
  & LMSYS    & 685,709 & 171,375 \\
\midrule
\multirow{3}{*}{Llama-3.1-8B}
  & ShareGPT & 24,117  & 6,047 \\
  & Alpaca   & 41,106  & 10,278 \\
  & LMSYS    & 656,083 & 164,014 \\
\midrule
\multirow{3}{*}{Qwen3-30B-A3B}
  & ShareGPT & 10,196  & 2,563 \\
  & Alpaca   & 40,639  & 10,155 \\
  & LMSYS    & 78,484  & 19,642 \\
\bottomrule
\end{tabular*}
\vspace{-0.5em}
\end{table}

\subsection{Model Details}
\label{app:model_details}

Our evaluation covers two model regimes. \textbf{Standard models} refer to general-purpose instruction-following LLMs without an explicit reasoning-oriented generation mode; this category includes \textbf{Llama-3.2-1B-Instruct} and \textbf{Llama-3.1-8B-Instruct}, both dense decoder-only Transformers. \textbf{Reasoning-capable models} refer to models whose responses may include explicit intermediate thinking content and longer deliberative trajectories; in our experiments this category is represented by \textbf{Qwen3-30B-A3B}, a mixture-of-experts (MoE) model.

We use this standard/reasoning distinction to characterize the target model family rather than the dataset. The same ShareGPT, Alpaca, and LMSYS prompts are used across models, while the output-length labels are regenerated separately for each target model. The reasoning-capable setting is more challenging for length prediction because generations are typically longer and more variable, and in multi-turn data the model may compress or drop earlier thinking content across turns. This is why, for Qwen3-30B-A3B, we retain only the first turn of ShareGPT and LMSYS when constructing labels, as described before.

\subsection{Baseline Descriptions}

\textbf{BERT-based Predictor.} We utilize \texttt{bigbird-roberta-base}~\footnote{\url{https://huggingface.co/google/bigbird-roberta-base}}~\cite{zaheer2021big} as the backbone for a regression model. Specifically, we take the representation of the \texttt{[CLS]} token and feed it into a two-layer fully connected regression head with ReLU activation, following the same design as in OUTLETS and the MLP baseline. The model is trained with Mean Squared Error (MSE) loss for 10 epochs, with a learning rate of $2 \times 10^{-5}$ and a batch size of 20.

\textbf{OPT-based Predictor.} We utilize \texttt{opt-125m}~\footnote{\url{https://huggingface.co/facebook/opt-125m}}~\cite{zhang2022opt} to treat length prediction as a classification task. The target length range [0, 2,048] is discretized into 50 uniform bins. A linear classification head is applied to the hidden state of the last valid token. The model is trained using Cross Entropy loss for 10 epochs with a learning rate of $2 \times 10^{-5}$.

\textbf{PIA-style Prompting Baseline.} We implement a lightweight PIA-style prompting baseline, where the LLM is asked to self-predict response length. We use a suffix prompting strategy where a specific instruction is appended to the user query. The instruction forces the model to output a length estimate on the first line before generating the actual response. \textbf{Note that due to time and resource constraints, we rely solely on this prompting mechanism without fine-tuning the model parameters.} We use greedy decoding (temperature $= 0.0$) to minimize variance. The specific prompt template is shown in Figure~\ref{fig:pia_prompt}.

\begin{figure}[t!]
    \centering
    \begin{tcolorbox}[colback=gray!10, colframe=gray!50, title=PIA Prompt Template]
        \textbf{\{Original User Prompt\}}
        
        Before responding to the above instruction, you have to predict the length of your response. Print the estimated number of words in your response in the first line. Then change to a new line to respond to the instruction.
    \end{tcolorbox}
    \caption{The prompt template used for PIA length prediction. The instruction is appended as a suffix to the user's original query.}
    \label{fig:pia_prompt}
\end{figure}

\section{System-Level Evaluation Details}
\label{app:system_eval}

To demonstrate practical utility, we integrate OUTLETS into a vLLM-based disaggregated serving system with one prefill instance and three decode instances. We evaluate performance on a filtered subset of the ShareGPT test set, selecting 1,914 requests with prompt lengths $\leq 50$ tokens. This filtering minimizes prefill-side interference, so that the measured gains primarily reflect improvements in decode-time scheduling rather than prompt-processing delays.

\begin{table*}[t!]
\centering
\caption{\textbf{Detailed System Performance at Saturation (100 QPS).} Comparison of the Baseline (RR + FCFS), RR + SJF, and LB + SJF. \textbf{Short}, \textbf{Medium}, and \textbf{Long} correspond to output lengths of $<800$, $[800, 1300)$, and $\geq 1300$ tokens, respectively. LB + SJF significantly reduces tail latency for short and medium tasks without starving long tasks, while achieving the highest overall system throughput.}
\label{tab:system_saturation}
\begin{tabular}{l|c|c|ccc}
\toprule
\multirow{2}{*}{\textbf{Method}} & \textbf{Throughput} & \textbf{Avg Latency}  & \multicolumn{3}{c}{\textbf{P99 Latency (s)}}  \\
 & (tokens/s) & (s)  & \textbf{Short} & \textbf{Medium} & \textbf{Long} \\
\midrule
RR + FCFS (Baseline) & 17,840.4 & 51.6  & 59.8 & 74.6 & 103.0  \\
RR + SJF & 18,108.2 & 50.8  & \textbf{38.0} & 61.7 & 100.8 \\
\textbf{LB + SJF} & \textbf{18,434.7} & \textbf{50.5}  & 39.0 & \textbf{60.1} & \textbf{100.1}  \\
\bottomrule
\end{tabular}
\end{table*}

\subsection{Compared Configurations}
To isolate the contribution of each scheduling component, we compare three configurations:
\begin{enumerate}
    \item \textbf{RR + FCFS (Baseline):} a standard vLLM (v0.13.0) backend using round-robin dispatch and FCFS admission, where requests are processed in arrival order;
    \item \textbf{RR + SJF:} round-robin dispatch combined with local SJF scheduling using the static length prediction from OUTLETS;
    \item \textbf{LB + SJF:} full instance-level load balancing, where each request is routed to the least-loaded decode instance, combined with local SJF scheduling.
\end{enumerate}

Following the saturated-regime scope in Section~\ref{sec:system_eval}, speculative acceleration is disabled in all configurations. This keeps the decoding service model fixed, so that differences in latency and throughput can be attributed to the routing and queue-ordering decisions driven by OUTLETS predictions.

\subsection{Results Under Saturation}
Table~\ref{tab:system_saturation} reports the detailed system performance under saturation (100 QPS). The results support three main conclusions.

\begin{enumerate}
    \item \textbf{SJF substantially mitigates HOL blocking.} The largest benefit appears in the tail latency of short tasks. Under the Baseline, short requests ($<800$ tokens) suffer from a P99 latency of 59.8s because they are frequently queued behind long-running generations. Once SJF is introduced, both RR + SJF and LB + SJF reduce this value to 38.0s and 39.0s, respectively. This corresponds to a 34.8\% reduction for short tasks and a 19.4\% reduction for medium tasks relative to the baseline. These results confirm that the predicted lengths are accurate enough to identify lightweight jobs and prioritize them effectively.

    \item \textbf{Load balancing improves cluster-wide throughput.} Although SJF alone improves latency, it does not fully address imbalance across decode instances. In the RR + SJF setting, requests are still assigned in a round-robin manner, so some workers may remain more congested than others. By adding instance-level LB, OUTLETS achieves the highest aggregate throughput of 18,434.7 tokens/s, compared with 17,840.4 tokens/s for the baseline, a 3.3\% improvement. This shows that predicted lengths are useful not only for local job ordering, but also for global request routing.

    \item \textbf{The gains for short jobs do not come at the expense of long jobs.} A common concern with SJF is starvation, namely that long jobs may be repeatedly delayed while short jobs keep arriving. In our experiments, this issue does not materialize. The P99 latency for long tasks ($\geq 1300$ tokens) under LB + SJF is 100.1s, slightly lower than 103.0s in the baseline. This suggests that the efficiency gains from better routing and reduced queue fragmentation are large enough to improve short-job responsiveness without harming heavy workloads.
\end{enumerate}

\subsection{Controlled comparison of predictors in the serving system}
\label{draft:predictor-system}

We evaluate how the choice of output-length predictor changes serving
performance under RR+SJF and LB+SJF.  Within each policy, all serving settings
are fixed and only the predictor changes.  The workload contains 1,914
requests with an average output length of approximately 1,127 tokens.  Each
complete serving experiment is repeated five times, and we report the mean.
The no-prediction setting is included as a reference.

Obtaining a stable Oracle requires additional care.  Greedy decoding
($\text{temperature}=0$) can still yield different outputs when dynamic batch
composition changes floating-point numerics \citep{he2025nondeterminism}.  We
therefore enable \texttt{VLLM\_BATCH\_INVARIANT=1}.  This mode adds performance
overhead, so absolute throughput should not be compared directly with the
original saturation table.  All predictors and the Oracle in the comparison
below use the same corresponding configuration, making within-table relative
comparisons controlled.

The reference output lengths are collected separately for the RR+SJF and
LB+SJF serving tests.  Because the realized generations can differ slightly
across tests even under greedy decoding, the same predictor has a slightly
different static MAE in the two tables.  We therefore report the MAE measured
against the output lengths realized in each corresponding serving test rather
than forcing a shared set of reference lengths across policies.

\begin{table*}[h]
\centering
\small
\setlength{\tabcolsep}{4.2pt}
\caption{Effect of the predictor under RR+SJF.  Serving metrics are means over
five complete runs.}
\label{tab:draft-predictor-rr}
\begin{tabular}{lrrrrrr}
\toprule
Predictor & Static MAE & Throughput & Avg. lat. & Short P99 & Medium P99 & Long P99 \\
& & (tok/s) & (s) & (s) & (s) & (s) \\
\midrule
No predictor & ---      & 12,018.34 & 77.61 & 91.66 & 119.82 & 166.53 \\
PIA          & 1,012.47 & 11,916.80 & 72.66 & 87.74 & 120.52 & 165.78 \\
BERT         & 308.10   & 12,045.03 & 70.83 & 46.25 & 115.17 & 159.05 \\
OPT          & 305.48   & \best{12,270.52} & 70.19 & 46.34 & 112.17 & 158.71 \\
MLP          & 265.41   & 12,100.33 & 69.70 & 46.28 & 110.92 & 160.42 \\
OUTLETS      & 219.77   & 12,260.99 & 68.92 & \best{46.07} & 105.92 & 156.91 \\
Oracle       & \best{0.00} & 12,197.95 & \best{68.19} & 46.14 & \best{95.65} & \best{153.65} \\
\bottomrule
\end{tabular}
\end{table*}

\begin{table*}[h]
\centering
\small
\setlength{\tabcolsep}{4.2pt}
\caption{Effect of the predictor under LB+SJF.  Serving metrics are means over
five complete runs.}
\label{tab:draft-predictor-lb}
\begin{tabular}{lrrrrrr}
\toprule
Predictor & Static MAE & Throughput & Avg. lat. & Short P99 & Medium P99 & Long P99 \\
& & (tok/s) & (s) & (s) & (s) & (s) \\
\midrule
No predictor & ---      & 12,156.05 & 77.54 & 90.85 & 120.81 & 165.73 \\
PIA          & 1,014.71 & 11,748.30 & 72.11 & 82.03 & 119.88 & 166.55 \\
BERT         & 309.44   & 11,933.24 & 70.59 & 45.92 & 113.98 & 161.00 \\
OPT          & 306.40   & 12,287.54 & 69.96 & 54.33 & 114.13 & 158.80 \\
MLP          & 267.98   & 12,341.39 & 69.01 & 45.42 & 110.74 & 156.45 \\
OUTLETS      & 221.92   & 12,363.31 & 68.71 & \best{45.28} & 104.12 & 154.55 \\
Oracle       & \best{0.00} & \best{12,422.32} & \best{67.70} & 45.46 & \best{97.25} & \best{150.33} \\
\bottomrule
\end{tabular}
\end{table*}

The results show the same broad trend under both routing policies: more
accurate output-length prediction generally lowers latency, and it usually
improves throughput when combined with prediction-based load balancing.
OUTLETS gives the strongest overall latency profile among non-oracle methods
and remains among the strongest in throughput.  The largest improvements over
weaker predictors occur for medium and long requests.

The relationship is not strictly monotonic for every metric or latency bucket.
This is expected because SJF depends on relative job ordering rather than
pointwise length accuracy.  Misordering jobs with substantially different
service times incurs more scheduling regret than exchanging similarly sized
jobs.  Once a predictor reliably separates very short requests from long
requests, further MAE reductions may have little effect on short-request P99
while still improving medium/long latency and prediction-based workload
estimation.  This interpretation is consistent with analyses of scheduling
with noisy service-time predictions, where imperfect estimates remain useful
when they preserve sufficient job ordering \citep{mitzenmacher2020price}.

\section{Ablation Studies}
\label{app:ablation}

\subsection{Impact of Joint Training}

The central design choice in OUTLETS is to train a shared speculative backbone for two objectives simultaneously: autoregressive drafting and remaining-length regression. This design is appealing from a systems perspective because it avoids introducing a separate auxiliary model. At the same time, it raises a natural concern: the two tasks operate at different semantic levels. Drafting is a local token-level prediction problem, while length prediction requires global trajectory awareness and an understanding of when generation is likely to terminate. If these objectives interfere strongly, joint training could degrade speculative decoding quality or weaken regression accuracy.

Our ablation results suggest that this concern is largely unfounded. Empirically, the shared backbone is expressive enough to support both objectives at once. The results point to a consistent picture: the drafting objective preserves local next-token fidelity, while the auxiliary regression head extracts global trajectory signals from the same latent space with little interference. In other words, OUTLETS behaves like a multi-task model in which the two tasks are related enough to share useful representations, yet distinct enough that one does not substantially damage the other.

\subsubsection{OUTLETS vs. SD-ONLY.}
\begin{figure*}[t]
  \centering
  \includegraphics[width=\textwidth]{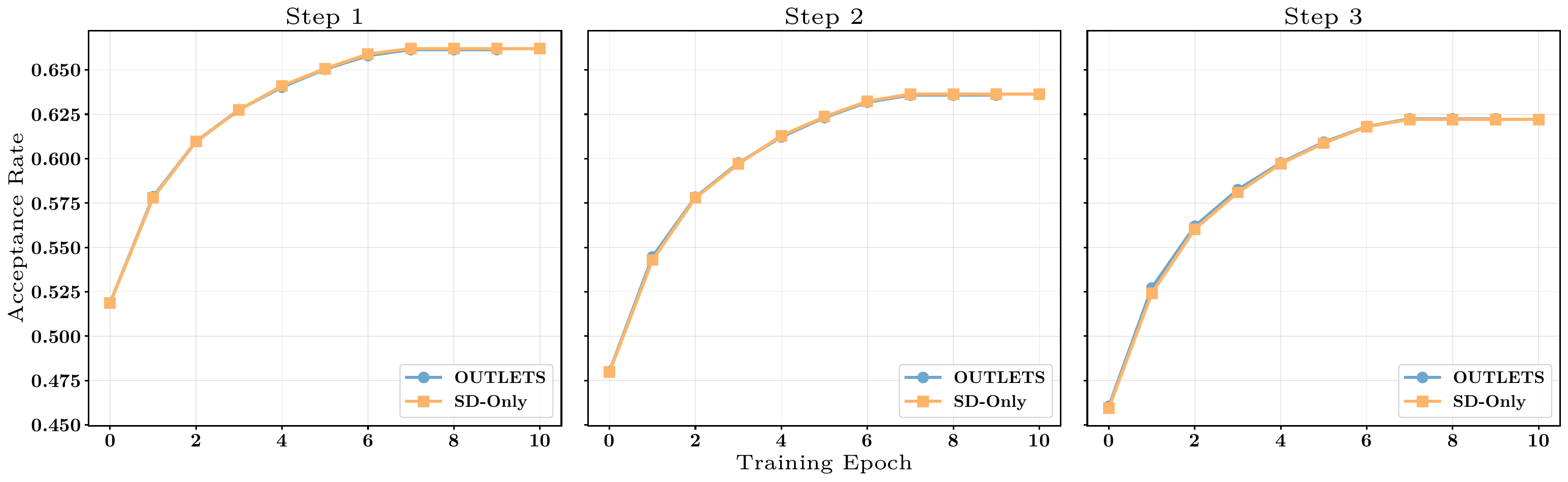}
  \caption{OUTLETS vs. SD-ONLY. Speculative acceptance rate across drafting steps on the test set. The acceptance rate is a standard proxy for speculative decoding quality: at each iteration, the draft model proposes $K$ tokens and the target model verifies them, accepting the longest matching prefix of length $m \le K$. We define the per-iteration acceptance rate as $\alpha = \frac{m}{K}$ and report its average over requests and iterations at each drafting step. As the speculative horizon increases, $\alpha$ typically decreases because mismatches compound over steps. In EAGLE-3, this degradation is already alleviated by the TTT mechanism; importantly, adding the auxiliary length-prediction objective in OUTLETS does not introduce an additional drop.}
  \label{fig:joint-vs-sdonly}
\end{figure*}

We first compare OUTLETS against an SD-ONLY baseline, i.e., a model trained only for speculative decoding without the auxiliary length-prediction objective. Figure~\ref{fig:joint-vs-sdonly} reports the speculative acceptance rate across drafting steps on the test set.
A key observation is that adding the regression objective does \emph{not} materially reduce the acceptance rate. The two curves remain very close throughout the drafting horizon, indicating that the shared backbone retains essentially the same drafting quality after joint training. This is an important result, because speculative acceptance is the primary indicator of whether the draft model continues to generate proposals that align with the target model. If the auxiliary task had introduced substantial gradient conflict, we would expect a visible degradation in acceptance, especially at deeper speculative steps where draft errors tend to accumulate. We do not observe such a trend.

This result supports our interpretation that the hidden space learned by the draft model contains enough capacity to encode both local and global information at the same time. The drafting head primarily relies on fine-grained local semantics needed for next-token prediction, whereas the regression head probes broader signals related to generation progress and eventual termination. Under this view, the length-prediction head behaves largely as a ``read-only'' consumer of the shared representation: it extracts trajectory information already present in the backbone rather than forcing the backbone to abandon the features needed for speculative decoding.

\subsubsection{OUTLETS vs. LP-ONLY.}
We further compare OUTLETS with an LP-ONLY baseline trained solely for length prediction. Unlike the SD-ONLY comparison, which tests whether joint training harms drafting, this comparison asks whether sharing the backbone with speculative decoding remains competitive with dedicating the full model to regression. Table~\ref{tab:joint-vs-lponly} reports the results.

A clear pattern emerges. In the \textit{static} setting, OUTLETS is generally better than or comparable to LP-ONLY, winning or tying in most configurations. In the \textit{dynamic} setting, LP-ONLY consistently achieves lower MAE, though the margin is modest.

\paragraph{Static prediction.}
Static prediction is data-sparse: each request provides only one target, the remaining length at prefill time. Under such limited supervision, LP-ONLY is more likely to overfit dataset-specific correlations. OUTLETS benefits from joint training because the drafting objective encourages the backbone to learn richer and more stable representations, which in turn regularize the regression head. This makes OUTLETS competitive with, and often better than, LP-ONLY in the static regime.

This advantage is especially relevant in practice, since static prediction is the setting directly used for pre-decode scheduling decisions such as dispatch and initial load balancing. OUTLETS therefore offers strong accuracy without requiring a separate predictor.

\paragraph{Dynamic prediction.}
Dynamic prediction provides much denser supervision, since every decoding step yields a remaining-length target. In this regime, LP-ONLY can devote all of its capacity to regression and specialize more aggressively, which explains its consistently lower MAE. OUTLETS, by contrast, must maintain representations that serve both drafting and regression, leading to a small but consistent specialization cost.

Still, this gap should be viewed in systems context. Although LP-ONLY performs better in dynamic prediction, the gain is modest relative to the deployment cost of an additional model. OUTLETS trades a small amount of accuracy for a simpler and more efficient design with lower memory overhead and tighter integration into the serving pipeline.

\begin{table*}[t!]
\caption{OUTLETS vs. LP-ONLY. We report MAE (lower is better). ``Avg. Len'' denotes the total output-length statistics for static prediction and the remaining-length target statistics aggregated over decoding steps for dynamic prediction ($\mu \pm \sigma$).}
\label{tab:joint-vs-lponly}
\centering
\begin{small}
\begin{tabular}{lll c cc}
\toprule
Model & Scenario & Dataset & Avg. Len & OUTLETS & LP-ONLY \\
\midrule
\multirow{6}{*}{Llama-3.2-1B} 
& \multirow{3}{*}{Static}
& ShareGPT & 324.4 $\pm$ 268.9 & 100.1 & \best{97.1} \\
& & Alpaca  & 237.6 $\pm$ 198.3 & \textbf{53.4} & 54.1 \\
& & LMSYS   & 246.6 $\pm$ 276.8 & \textbf{69.6} & 71.7 \\
\cmidrule(l){2-6}
& \multirow{3}{*}{Dynamic}
& ShareGPT & 274.2 $\pm$ 226.7 & 63.8 & \textbf{60.3} \\
& & Alpaca  & 202.1 $\pm$ 175.8 & 31.8 & \textbf{28.9} \\
& & LMSYS   & 279.1 $\pm$ 256.7 & 41.7 & \textbf{39.4} \\
\midrule
\multirow{6}{*}{Llama-3.1-8B} 
& \multirow{3}{*}{Static}
& ShareGPT & 355.9 $\pm$ 276.7 & \textbf{80.6} & 81.1 \\
& & Alpaca  & 250.3 $\pm$ 206.9 & \textbf{46.0} & \textbf{46.0} \\
& & LMSYS   & 257.6 $\pm$ 299.0 & \textbf{70.6} & 73.2 \\
\cmidrule(l){2-6}
& \multirow{3}{*}{Dynamic}
& ShareGPT & 286.0 $\pm$ 232.9 & 67.6 & \textbf{65.3} \\
& & Alpaca  & 211.2 $\pm$ 174.7 & 39.4 & \textbf{37.0} \\
& & LMSYS   & 302.9 $\pm$ 292.7 & 38.8 & \textbf{35.8} \\
\midrule
\multirow{6}{*}{Qwen3-30B-A3B} 
& \multirow{3}{*}{Static}
& ShareGPT & 1053.9 $\pm$ 536.8 & \textbf{186.7} & 195.0 \\
& & Alpaca  & 725.9 $\pm$ 374.2 & \textbf{172.1} & \textbf{172.1} \\
& & LMSYS   & 791.7 $\pm$ 536.1 & 166.8 & \textbf{164.9} \\
\cmidrule(l){2-6}
& \multirow{3}{*}{Dynamic}
& ShareGPT & 664.1 $\pm$ 460.8 & 117.1 & \textbf{103.3} \\
& & Alpaca  & 459.9 $\pm$ 353.9 & 94.0 & \textbf{86.6} \\
& & LMSYS   & 578.8 $\pm$ 441.2 & 103.2 & \textbf{92.0} \\
\bottomrule
\end{tabular}
\end{small}
\vskip -0.1in
\end{table*}

\subsection{Feature Importance Analysis}

We ablate the main components of the EAGLE-3-style backbone, including multi-level feature fusion, input embeddings, TTT, and the draft decoder layers, to understand which parts are most important for length prediction. Among them, the draft decoder has the largest impact: removing it causes a substantial increase in MAE, showing that fused hidden states alone are insufficient.

This suggests that accurate length prediction depends not only on the information contained in the hidden states, but also on how that information is processed. The draft decoder appears to transform static fused features into trajectory-aware representations that better capture progress toward termination. By contrast, feature fusion, input embeddings, and TTT provide smaller gains. Overall, these results support our central claim that OUTLETS benefits from reusing the speculative backbone as a forecasting module, with the draft decoder serving as its key component.

\subsection{Regression vs. Classification}

We also compare two formulations for output-length prediction: direct regression and bucket-based classification. Direct regression consistently performs better. This is expected because output length is an ordered quantity, and regression preserves the numerical distance between predictions and targets, which is better aligned with MAE.

Classification introduces discretization artifacts: lengths within the same bucket are treated as identical, while lengths near a bucket boundary may be assigned different labels despite being numerically close. This weakens the supervision signal and makes learning less stable. Regression also provides a more natural interface for downstream scheduling, since serving systems typically require a scalar estimate rather than a bucket label. These results therefore support our choice of a direct regression head.

\subsection{Cross-Domain Generalization}
\label{draft:cross-domain}

To evaluate whether OUTLETS is restricted to chat-style requests, we train a
new predictor only on LMSYS and increase the maximum sequence length from
2,048 to 4,096 tokens on Qwen3-30B-A3B model.  We then evaluate it directly on GSM8K
\citep{cobbe2021gsm8k} and HumanEval \citep{chen2021humaneval}, without using
mathematical-reasoning or code-generation data during training.

\begin{table}[h]
\centering
\small
\caption{Zero-shot cross-domain output-length prediction.  The OUTLETS model
is trained only on LMSYS with a maximum sequence length of 4,096 tokens.}
\label{tab:draft-cross-domain}
\begin{tabular}{lrrr}
\toprule
Test dataset & Avg. len & Std. & MAE \\
\midrule
GSM8K    & 1,561.24 & 756.12 & 397.9 \\
HumanEval & 2,324.56 & 825.50 & 506.6 \\
\bottomrule
\end{tabular}
\end{table}

The two datasets contain substantially longer and more variable generations than the original chat workloads. Nevertheless, the MAEs are only 25.5\% and 21.8\% of the corresponding average output lengths on GSM8K and HumanEval, respectively. This behavior is consistent with the log-space regression objective, which emphasizes relative rather than absolute length error. Due to computational constraints, we evaluate only OUTLETS in this cross-domain experiment. The results provide preliminary evidence that the learned signal transfers to longer mathematical-reasoning and code-generation trajectories.

\end{document}